\documentclass{article}

\usepackage{arxiv}
\usepackage[]{algorithm2e}
\usepackage{neuralnetwork}
\usepackage[utf8]{inputenc} 
\usepackage[T1]{fontenc}    
\usepackage{hyperref}       
\usepackage{url}            
\usepackage{booktabs}       
\usepackage{amsfonts}       
\usepackage{nicefrac}       
\usepackage{microtype}      
\usepackage{lipsum}
\usepackage{amsmath}
\usepackage{amssymb}
\usepackage{amstext}
\usepackage{amsthm}
\usepackage{chemarrow}
\usepackage{mathtools}

\usepackage{romannum}
\usepackage{tikz}
\usetikzlibrary{arrows,positioning,trees} 
\usepackage{subcaption}

\title{Gradient-based Causal Structure Learning with Normalizing Flow}

\author{
 Xiongren Chen \\
  School of Computer Science\\
  University of Adelaide\\
  \texttt{xiongren.chen@adelaide.edu.au} \\
}

\begin{document}
\maketitle

\begin{abstract}
In this paper, we propose a score-based normalizing flow method called DAG-NF to learn dependencies of input observation data. Inspired by Grad-CAM in computer vision, we use jacobian matrix of output on input as causal relationships and this method can be generalized to any neural networks especially for flow-based generative neural networks such as Masked Autoregressive Flow(MAF) and Continuous Normalizing Flow(CNF) which compute the log likelihood loss and divergence of distribution of input data and target distribution. This method extends NOTEARS which enforces a important acylicity constraint on continuous adjacency matrix of graph nodes and significantly reduce the computational complexity of search space of graph. 
\end{abstract}


\section{Introduction}
How to find meaningful relationships, especially causal relationships, from massive amounts of non-sequence observation data is one of the research areas most likely to create business value and make scientific discoveries in data science, and is receiving widespread attention from international peers. Causality strictly distinguishes between the cause and effect variables, and has an important role that cannot be replaced by the relationship in revealing the mechanism of things and guiding the intervention behavior.
Causality learning has been wildly studied in many applications, for example, \cite{Sanford2011} applied causal structure into operational risk model to learn which human factor attributing to the operational risk in finance, while in medicine field, \cite{Subramani2000} learned the causal structure of clinical conditions and outcomes  from static observation data and the causal network of protein interaction from the science paper \cite{sachs2005} has been commonly accepted by researchers in this filed. There are also studies of causal inference in epidemiology\cite{Miguel2000}, education \cite{Rajeev1999} and environmental health\cite{Jundong2014}.

In general, randomized experiments is a effective method for obtaining causal relationships\cite{Pearl2009}, but random experimentation is mostly impossible to intervene or the cost of interventions are enormous. For example, if we need to analyze the causal relationships between interest rates and stock market equity factors, we can't shift the interest rate and then observe the fluctuations of equity factors in the stock market, as the resulting impact is huge. So we have to rely on observational data. 

The existing methods of causal inference from observed data can be broadly classified into three categories, which are constraint-based, score-based and structural causal model based methods. While Constraint-based methods test the specified structure by conditional independent test which can be implemented by statistical information measures, A score-based approach uses a scoring function to quantify how well a Bayesian network fits a given distribution of data and then uses a search algorithm to find the graph structure that best fits the data. structural causal model based methods describes the mechanism for generating data between causes and effects variables.

In this paper, we propose a score-based normalizing flow method called DAG-NF to learn dependencies of input observation data. Inspired by Grad-CAM\cite{cam2017} in computer vision, we use jacobian matrix of output w.r.t. input as causal relationships and this method can be generalized to any neural networks especially for flow-based generative neural networks such as Masked Autoregressive Flow(MAF)\cite{maf2017} which compute the log likelihood loss and divergence of distribution of input data and target distribution. This method extends NOTEARS\cite{zheng2018} which enforces a important acyclicity constraint on continuous adjacency matrix of graph nodes and significantly reduce the computational complexity of search space of graph. 

The first contribution of our work in this chapter is that we provide a general framework of neural networks to causal structure learning on non-sequence data. This neural networks can be any machine learning architectures that can compute the jacobian matrix of output on input, for example, a normal MLP neural network or generative models such as MAF\cite{maf2017}. Similar to what machine learning is doing, this framework allows for alter the structure or parameters of neural networks to achieve better results. The second contribution is that we design a new architecture with self-masking to obtain causal relationships of input variables. The last contribution is that we did variety experiments comparing state-of-art causal inference methods and it shows that our performance is competitive in all experiments but our method is more flexible because it only requires neural networks to compute the jacobian matrix. 

\section{Background and Related Work}
\label{sec:headings}

\subsection{Graphical Causal Models and Directed Acyclic Graph(DAG)}
A graphical causal model is a formalism for representing causal relations. A graphical causal model $\mathbb{G}(V,E) $ includes (1) $X \in \mathbb{R}^{d}$ denotes a set of random variables with $d$ dimensions while $V_i$ in graph denotes each node in $X$ and (2) a set of directed edges $E$ denotes the causes and effects between pairs of nodes in graph $G$ with the meaning of the "parent" cause node pointing the effect node, and (3) a joint probability distribution$P_X$ fits data of all the random variables, and (4) we assume that there are no cycles(no closed directed paths) or feedbacks among the edges in a graphical causal model as it allows simple interpretations. Therefore, we commonly use Directed Acyclic Graph(DAG) instead of Graphical Causal Models in this paper. The distribution $P_X$ as $P(x)$ follows Markov property on $\mathbb{G}(V,E) $ and can be decomposed to a product of simple distributions,
\begin{equation} \label{eq:3} P(x) = \prod_d P(x_j | x_{\pi_j^G}) \end{equation} 
where $\pi_j^G$ denotes the set of parents of node $j$ in $G$ and $x_{\pi_j^G}$ denotes the random vector containing the variables corresponding to the parents of $j$ in $G$

\subsection{Structural Causal Models(SCMs)}
A structural causal model(we sometimes call structural equation model) usually has $d$ assignments,
\begin{equation} \label{eq:4} X_j := f_j(x_{\pi_j^G}, N_j) \end{equation} 
where $\pi_j^G$ denotes the set of parents of node $j$ in $G$ and $x_{\pi_j^G}$ denotes the random vector containing the variables corresponding to the parents of $j$ in $G$, $f_j$ is a linear or nonlinear function satisfying some mild regularity conditions and a product distribution of the noises $N_j$ is joint independent. In a SCM, all the variables can be computed by the parent node variables in the $G$. A SCM will define a unique distribution $P_X$ over X in a DAG\cite{peter2014}. 

\subsection{Structure Identifiability}
Some SCMs with corresponding graphs can refer to a same data distribution $P_X$\cite{peter2014} and we sometimes should require assumptions to restrict the functions $f_j$ to find a identifiablity result. For example, in an Additive Noise Model(ANM), we have structural assignments as the form of
\begin{equation} \label{eu_eqn} X_j := f_j(x_{\pi_j^G}) + N_j,\hspace{1cm} j=1,...d, \end{equation} 

The function $f_j$ in ANMs rely on the parameters and is is not constant, which ensure the causal minimality\cite{Peter2017} but the restriction of function $f_j$ can not obtain full structure idenfiability. However, there are some examples extend the framework of ANMs obtain the uniqueness, for example, the linear Gaussian case with equal error variances, the linear non-Gaussian ANMs and nonlinear Gaussian ANMs\cite{Peter2017}. In this paper, we assume the nonlinear Gaussian cases and use special neural network to fit the function $f_j$ to make sure the requirement of nonlinear satisfied.

\subsection{Score-based methods and NOTEAR}
Scrore-based methods test a set graph structures seeking to find a graph with the highest score to fit the data and this optimization problem has the form of,
\begin{equation} \label{eu_eqn} \hat{G} :=  \arg\max_{DAG} S(D,G)
\end{equation}
where $S(D,G)$ is a score function over given data and a graph structure and $\hat{G}$ is the optimal dag with highest score. There are some popular score function, for example, Bayesian information criterion (BIC) scores\cite{hau1988} and Bayesian Dirichlet(BD) score\cite{geiger1994}. The search space grows superexponentially with the dimension $d$ of the data and it is a NP-hard optimization problem to solve\cite{Chickering1996}. However, the greedy search algorithms\cite{Chickering2002}, order search\cite{sachs2005} and coordinate descent\cite{gu2018} techniques can be applied to find the optimal highest score among the candidate graphs. For ANMs, method of regression with subsequent independence test(RESIT) is proposed to linear ANMs\cite{peter2014} and ICA\cite{comon1994} is appled to solve the nonlinear ones\cite{shimizu2006}. The greedy search can often guarantee a optimal solution but the scale of number of variables is often limited due to the huge search space.

NOTEAR\cite{zheng2018} reformulates the combinatorial optimization problem into a continuous problem with acyclicity constraint and significantly reduces the size of the search space of linear structure equation models. For a linear SEM with the form of,
\begin{equation} 
X_j = W^TX + N_j,\hspace{1cm} j=1,...d,
\end{equation}
where $W \in \mathbb{R}^{m \times m}$ denotes weighted adjacency matrix of $G$. $W$ is a DAG if and only if\cite{zheng2018},

\begin{equation}  \label{eq:1}
h(W) = tr(e^{W\circ W}) - d = 0
\end{equation}
or\cite{yu2019},
\begin{equation}  \label{eq:2}
h(W) = tr[(I+\alpha W\circ W)^d] - d = 0
\end{equation}
where $\circ$ is for Hadamard product, $tr$ is for trace function of matrix, $\alpha$ can be any value greater than 0 and $e^W$ is the matrix exponential of W. It is simple to compute the gradient,
\begin{equation} 
\nabla  h(W) = (e^{W\circ W})^T \circ 2W 
\end{equation}

\subsection{Related Work}
Most existing methods of causal inference are constraint-based, score-based and structure causal model based methods. Score-based methods which we use in the chapter define a score function such as Bayesian information criterion (BIC) scores\cite{hau1988} and Bayesian Dirichlet(BD) score\cite{geiger1994} and try to optimal one with highest score from a set of DAGs. Due to the huge superexponential search space with growing number of variables, greedy search algorithm is introduced to solve this intractable problem. GES\cite{Chickering2002} use BIC as score function and try to find the local optimal graph from adding edges and removing edges phases. For SEMs, representative algorithms include Linear Non-Gaussian Acyclic Model(LiNGAM\cite{shimizu2006}), Post-NonLinear(PNL\cite{zhang2009,zhang2006}), ANMs\cite{peter2011} and their extensions\cite{hoyer2009} and Information Geometric Causal Inference(IGCI\cite{dan2010}). LiNGAM\cite{shimizu2006} assumes the function $f_i$ is linear, non-Gaussian noises and acyclic dependency paths, which is based on Independent Component Analysis(ICA) and rely heavy on the initial solutions. In PNL\cite{zhang2009,zhang2006}, there are two non-linear functions in an assignment as the form of $ X_j :=f_{j2}(f_{j1}(x_{\pi_j^G}) + N_j) $. PNL has broad and general applicability but the two non-linear functions increased computational complexity. ANMs\cite{peter2011,hoyer2009} describe a method for implementing the discovery of causality between binary variables under nonlinear conditions and their extensions\cite{peter2014} extend the ANM model to the case of multidimensional variables with the method of regression with subsequent independence test(RESIT), which is applicable to cases of same variance error data or discrete data. IGCI\cite{dan2010} assumes that the causal influence process is noiseless and the derivatives of nonlinear functions between two variables are statistically uncorrelated. Therefore, ICGI-like methods focus primarily on no-noise or low-noise and complex functions cases. Causal Additive Model(CAM\cite{cam2014}) is a example of nonlinear Guassian ANM, which satisfies the requirement of idenfiablity, having a form of $ X_j := \sum_{k\in {PA_j} } f_{j,k}(X_k) + N_j$, where $PA_j$ denotes the parent nodes of $X_j$ in $G$. However, such assumptions of functions are too strong to be generalizable. NOTEAR\cite{zheng2018} reformulates the combinatorial optimization problem into a continuous problem with acyclicity constraint and significantly reduces the size of the search space of linear structure equation models. DAG-GNN\cite{yu2019} extend NOTEAR's continuous linear Structural Equation Model(SEM) to non-linear model with VAE and graph convolutional neural network(GCNN), learning a neural network by maximizing an evidence lower band. SAM model\cite{Kalainathan2019} applied GAN to learning data structure with acyclicity constraint. Recently, gradient-based methods are proposed by GraN-Dag\cite{gran2020}, Masked Gradient-Based Causal Structure Learning(Masked-Grad\cite{mask2020}) and NOTEAR's author's new paper Learning Sparse Nonparametric DAGs(Sparse-DAG\cite{sparse2020}), determining the causal relationship between two variables through neural network connectivity. Gradient-based methods outperform DAG-GNN at all aspects of benchmarks in empirical comparisons, which are proven to be good way to learn causal structure. However, GraN-Dag use weights in neural networks to ensure connectivity with less generalizability compared to jacobian matrix which we use in this paper and the loss function does not include the divergence of distribution of input data and target data. Masked-Grad try to learn a binary matrix instead of continuous weighted matrix with the framework of GraN-Dag. Sparse-DAG ensure the connectivity of input layer and first hidden layer but not for all layers of neural network, however, its performance is competitive with less sample of data. 

\section{DAG Structure Learning with Jacobian Matrix}
\label{sec:headings}

\subsection{Jacobian Matrix as Causal Dependencies}
Suppose $x_j := f_j(x_k) $ is nonlinear function with first-order partial derivatives exist $\frac{\partial f_j}{\partial x_k }$ on $\mathbb{R}^{d}$, we can define a jacobian matrix of a SEM with the form of $X_j := f_j(x_{\pi_j^G}, N_j)$ over random variables $X$ as,
\begin{equation} \label{eu_eqn} J =
\begin{bmatrix} \frac{\partial f}{\partial x_1} \cdot \cdot \cdot \frac{\partial f}{\partial x_d}
\end{bmatrix} = 
\left[
\begin{array}{ccc}
  \frac{\partial f_1}{\partial x_1} & \cdots & \frac{\partial f_1}{\partial x_d} \\
   \vdots & \ddots & \vdots \\
   \frac{\partial f_d}{\partial x_1} & \cdots & \frac{\partial f_d}{\partial x_d}
\end{array}
\right]
\end{equation} 
Sparce-Dag\cite{sparse2020} proposed if $||\frac{\partial f_j}{\partial x_k} ||_{L^2} = 0$ then there is no dependency of $x_j$ on $x_k$ ,where $||\cdot||_{L^2}$ is the usual $L^2$ norm. Therefore, the weighted adjacency matrix of nonlinear extension can be defined as 
$ W(f) = || J ||_{L^2} $
and the extension of nonlinear case of \ref{eq:1} or \ref{eq:2} would be $ h(W(f)) = 0$ 

\subsection{Conditional Independent Score Function}
For equation (\ref{eq:3}) , we try to decompose $P_X$ as a product of some simple known distribution such as Gaussian and have restricts to function $f_j$ so that idenfibility can be satisfied. We can start from assighment funcitons of SEMs to obtain the detail transformation. We can have the form of $N_j$ from equation (\ref{eq:4}) as,
\begin{equation} N_j := f^{-1}_j(x_{\pi_j^G}, x_j) \end{equation}
where $N_j$ is independent variable and it has same dimension $d$ of input $X$, following the decomposition,
\begin{equation}  q(N) = \prod_d q(N_j) \end{equation}
Where q() can be any simple known distribution such as Gaussian distribution. According to the rules of change of variables,
\begin{equation}  P_X = p(x) = q(f^{-1}(x)) \left|det(\frac{\partial f^{-1}}{\partial x}) \right| \end{equation}
where $\det(\frac{\partial f^{-1}}{\partial x})$ is determinant of Jacobian matirx of $f^{-1}$ over $x$. Now we have the form of maximum log likelihood loss function would change from,
\begin{equation} 
\max_{\phi}\mathbb{E}_{X \sim P_X}\sum_{j=1}^d \log p_j(X_j \mid  X_{\pi_j^\phi};\phi_{(j)}) 
\end{equation}to,
\begin{equation} 
-Loss = \max_{\theta}\mathbb{E}_{X \sim P_X} \left[ \sum_{j=1}^d \log q(N_j) + \log \left|\det(\frac{\partial f^{-1}}{\partial x}) \right| \right]
\end{equation}
where $G_\phi$ is the optimal solution of graph with parameters $\phi$ and $\theta$ is the set of parameters of nonlinear function such as neural networks.
\subsection{Masked Autoregressive Density Estimation(MADE)}
In MADE\cite{Germain2015}, the joint distribution can be decomposed into a product of
one-dimensional conditionals as $q(x) =\prod_i q(x_i | x_{1:i-1}) $ which called autoregressive property. MADE design a masked matrix in each layer of neural networks to ensure the output $x_i$ depend only on the preceding input $x_{1:i-1}$. For example, a neural network of input with 3 nodes can be designed as,

\begin{neuralnetwork}[height=4]
        \newcommand{\x}[2]{$x_#2$}
        \newcommand{\y}[2]{$\hat{x}_#2$}
        \newcommand{\hfirst}[2]{\small $#2$}
        \newcommand{\hsecond}[2]{\small $h^{(2)}_#2$}		
        \newcommand{\nodetextxnb}[2]{\ifnum #2<3 1 \else 2 \fi}
        \newcommand{\mynodetext}[2] {
		}
        \inputlayer[count=3, bias=false, title=Input\\layer, text=\x]
        \hiddenlayer[count=4, bias=false, title=Hidden\\layer 1, text=\nodetextxnb]
        	\link[from layer=0, to layer=1, from node=1, to node=1]
        	\link[from layer=0, to layer=1, from node=1, to node=2]
        	\link[from layer=0, to layer=1, from node=1, to node=3]
        	\link[from layer=0, to layer=1, from node=1, to node=4]
        	\link[from layer=0, to layer=1, from node=2, to node=3]
        	\link[from layer=0, to layer=1, from node=2, to node=4]

        \hiddenlayer[count=4, bias=false, title=Hidden\\layer 2, text=\nodetextxnb]
        	\link[from layer=1, to layer=2, from node=1, to node=1]
        	\link[from layer=1, to layer=2, from node=1, to node=2]
        	\link[from layer=1, to layer=2, from node=1, to node=3]
        	\link[from layer=1, to layer=2, from node=1, to node=4]
        	\link[from layer=1, to layer=2, from node=2, to node=1]
        	\link[from layer=1, to layer=2, from node=2, to node=2]
        	\link[from layer=1, to layer=2, from node=2, to node=3]
        	\link[from layer=1, to layer=2, from node=2, to node=4]
        	\link[from layer=1, to layer=2, from node=3, to node=3]
        	\link[from layer=1, to layer=2, from node=3, to node=4]
        	\link[from layer=1, to layer=2, from node=4, to node=3]
        	\link[from layer=1, to layer=2, from node=4, to node=4]
        \outputlayer[count=3, title=Output\\layer, text=\y] 
        	\link[from layer=2, to layer=3, from node=1, to node=2]
        	\link[from layer=2, to layer=3, from node=1, to node=3]
        	\link[from layer=2, to layer=3, from node=2, to node=2]
        	\link[from layer=2, to layer=3, from node=2, to node=3]
        	\link[from layer=2, to layer=3, from node=3, to node=3]
        	\link[from layer=2, to layer=3, from node=4, to node=3]
    \end{neuralnetwork}

In SEM language, the relationships of input X can be entailed by,
\begin{equation} 
\begin{split}
X_1 &:= N_1 \\
X_2 &:= f_2(X_1) + N_2 \\
X_3 &:= f_3(X_1,X_2) + N_3
\end{split}
\end{equation}

\subsection{MAF and Causal Structure Learning}
We can stack some blocks of MADE to form a Masked Autoregressive Normalizing Flow(MAF\cite{maf2017}). In MADE, $q(x) =\prod_i q(x_i | x_{1:i-1})$ and $q(x_i | x_{1:i-1})$ which can be a simple known distribution such as Gaussian parameterized by mean and variance,
\begin{equation}
    q(x_i \mid x_{1:i-1}) = \mathcal{N}(x_i \mid \mu_i,\,(\exp\alpha_i)^{2})
\end{equation}
Where $\mu_i = f_{\mu_i}(x_{1:i-1})$ and $\alpha_i = f_{\alpha_i}(x_{1:i-1})$ and both are nonlinear functions which can be fitted by neural networks. Therefore, The form of $f_j$ and $f^{-1}_j$ in MADE is given by,
\begin{equation} 
\begin{split}
f_j \implies X_j &= N_j\exp\alpha_j + \mu_j  \\
f^{-1}_j \implies N_j &= (X_j - \mu_j)\exp(-\alpha_j) \\
\end{split}
\end{equation}
where $N_j$ follows a normal Gaussian in here. And due to the autoregressive property, we can calculate determinant of the jacobian of $f^{-1}$ as the sum of triangular elements in the jacobian matrix. However, the transform of MADE rely heavily on the order of input variables($x_i$ only learned from previous variables and ignore the following variable) and it is too weak to fit a complex function and we have to combine more transform of MADE as a flow make it able to learn complex transformation. The structure is designed as follow,
\begin{align*}
X = h^{(0)}\; \autorightleftharpoons{\scriptsize$f_1$}{\scriptsize$f_1^{-1}$} h^{(1)}
\autorightleftharpoons{\scriptsize$f_2$}{\scriptsize$f_2^{-1}$} h^{(2)}
...
\autorightleftharpoons{\scriptsize$f_{n-1}$}{\scriptsize$f_{n-1}^{-1}$} h^{(n-1)}
\autorightleftharpoons{\scriptsize$f_{n}$}{\scriptsize$f_{n}^{-1}$} h^{(n)} = N
\end{align*}

We can easily check whether MAF can fit causal structure learning problem. We suppose that the causal order of the input variables is unknown, therefore, we connect 2 or more blocks of MADE formed as a flow to solve this problem. The first block of the flow can infer the relations of $x_i$ on the preceding variables $x_{1:i-1}$ while another one inferring relations of $x_i$ on the following variables $x_{i+1:d}$. For example, the relationships of input data $X$ in SEM language is,
\begin{equation} 
\begin{split}
X_1 &:= N_1 \\
X_2 &:= f_2(X_1) + N_2 \\
X_3 &:= f_3(X_1,X_2) + N_3
\end{split}
\end{equation}

And the order of input X is unknown. We assume the input order(not the causal order) of input X as ($X_2$,$X_1$,$X_3$) and ideally the network would be learned as follows,
         
\begin{neuralnetwork}[height=4, layerspacing=18mm]
        \newcommand{\x}[2]{\ifnum #2=1 ${x}_2$ \else \ifnum #2=2 ${x}_1$ \else ${x}_3$  \fi  \fi}
        \newcommand{\y}[2]{\ifnum #2=1 $\hat{x}_2$ \else \ifnum #2=2 $\hat{x}_1$ \else $\hat{x}_3$  \fi  \fi}
        \newcommand{\z}[2]{\ifnum #2=1 $\hat{x}_3$ \else \ifnum #2=2 $\hat{x}_1$ \else $\hat{x}_2$  \fi  \fi}
        \newcommand{\hfirst}[2]{\small $#2$}
        \newcommand{\hsecond}[2]{\small $h^{(2)}_#2$}		
        \newcommand{\nodetextxnb}[2]{\ifnum #2<3 1 \else 2 \fi}
        \newcommand{\mynodetext}[2] {
		}
        \inputlayer[count=3, bias=false, title=Input\\layer, text=\x]
        \hiddenlayer[count=4, bias=false, title=Hidden\\layer 1, text=\nodetextxnb]
        	\link[from layer=0, to layer=1, from node=1, to node=1]
        	\link[from layer=0, to layer=1, from node=1, to node=2]
        	\link[from layer=0, to layer=1, from node=2, to node=3]
        	\link[from layer=0, to layer=1, from node=2, to node=4]

        \hiddenlayer[count=4, bias=false, title=Hidden\\layer 2, text=\nodetextxnb]
        	\link[from layer=1, to layer=2, from node=1, to node=1]
        	\link[from layer=1, to layer=2, from node=1, to node=2]
        	\link[from layer=1, to layer=2, from node=2, to node=1]
        	\link[from layer=1, to layer=2, from node=2, to node=2]
        	\link[from layer=1, to layer=2, from node=3, to node=3]
        	\link[from layer=1, to layer=2, from node=3, to node=4]
        	\link[from layer=1, to layer=2, from node=4, to node=3]
        	\link[from layer=1, to layer=2, from node=4, to node=4]
        \outputlayer[count=3, title=Output\\layer, text=\y] 
        	\link[from layer=2, to layer=3, from node=1, to node=3]
        	\link[from layer=2, to layer=3, from node=2, to node=3]
        	\link[from layer=2, to layer=3, from node=3, to node=3]
        	\link[from layer=2, to layer=3, from node=4, to node=3]
        
        \inputlayer[count=3, bias=false, title=Input\\layer, text=\z]
        	\link[from layer=3, to layer=4, from node=1, to node=3]
        	\link[from layer=3, to layer=4, from node=2, to node=2]
        	\link[from layer=3, to layer=4, from node=3, to node=1]
        	
        \hiddenlayer[count=4, bias=false, title=Hidden\\layer 1, text=\nodetextxnb]
        	\link[from layer=4, to layer=5, from node=1, to node=1]
        	\link[from layer=4, to layer=5, from node=1, to node=2]
        	\link[from layer=4, to layer=5, from node=2, to node=3]
        	\link[from layer=4, to layer=5, from node=2, to node=4]
        \hiddenlayer[count=4, bias=false, title=Hidden\\layer 2, text=\nodetextxnb]
        	\link[from layer=5, to layer=6, from node=1, to node=1]
        	\link[from layer=5, to layer=6, from node=1, to node=2]
        	\link[from layer=5, to layer=6, from node=2, to node=1]
        	\link[from layer=5, to layer=6, from node=2, to node=2]
        	\link[from layer=5, to layer=6, from node=3, to node=3]
        	\link[from layer=5, to layer=6, from node=3, to node=4]
        	\link[from layer=5, to layer=6, from node=4, to node=3]
        	\link[from layer=5, to layer=6, from node=4, to node=4]
        \outputlayer[count=3, title=Output\\layer, text=\z] 
        	\link[from layer=6, to layer=7, from node=3, to node=3]
        	\link[from layer=6, to layer=7, from node=4, to node=3]
    \end{neuralnetwork}

\subsection{Add Acylicity Constraint to Loss function with Augmented Lagrangian}
The optimization problem is given by,
\begin{equation} 
\begin{split}
\min_{\theta} f(\theta) &= Loss \\
\textrm{s.t.} \; h(W(f)) &= 0
\end{split}
\label{eq:unconstrained}
\end{equation}
Where $Loss$ is the objective function and $h(W(f))$ is constraint and we can add the constraint to objective function to form a Lagrangian function,
\begin{equation} 
L_c(\theta,\lambda) = f(\theta) + \lambda h(W(f))
\end{equation}
Where $\lambda$ is the Lagrangian multiplier. The Lagrangian function is an unconstrained optimization problem and its solution is an optimal solution to the constrained problem (\ref{eq:unconstrained}). However, the Lagrangian function can not guarantee an optimal solution so we should add an augmented term the ensure the feasibility and optimal solution of the method, having the form of,
\begin{equation} 
L_c(\theta,\lambda) = f(\theta) + \lambda h(W(f)) + \frac{\rho}{2}\mid h(W(f)) \mid^2
\end{equation}
Where $\rho$ is the penalty parameter. The Augmented Lagrangian function is different from a normal Lagrangian function due to the penalty function. For the Augmented Lagrangian function, we increase penalty parameter until a large value for each iteration of unconstrained optimization. When penalty parameter approaches to $+\inf$, the Lagrangian function should satisfy $h(W(f))=0$. Since the penalty parameter is a finite enumeration of values, the result must be a locally optimal solution. During each iteration, the Lagrangian multiplier and the penalty parameter remain fixed. After the iteration, the Lagrange multiplier is updated according to the optimization results of the iteration by,
\begin{equation} 
\lambda^{k+1} = \lambda^{k} + \rho^k h(W(f))^k
\end{equation}
Where $k$ is for the $k$-iteration. For the update of penalty parameter, our strategy is progressively increase $\rho$ by\cite{yu2019},
$$
 \rho^{k+1} = 
  \begin{cases} 
   10 * \rho^k ,& \text{if } \mid h(W(f))\mid^k >\frac{1}{4} \mid h(W(f))\mid^{k-1} \\
   \rho^{k}       ,& \text{otherwise, }
  \end{cases}
$$
\section{Experiments}
In this section, we do experiments on synthetic data and real data to study the performance of Jacobian(gradient)-based methods specifying by neural networks of MAF. For DAG-MAF, we combine 6 blocks of MADE neural network with 1 hidden layers of 100 neurons and use Relu as activation function.

\textbf{Baselines} There are many nice works in this filed and we would like to compare our work to following works with the best performance as baselines:  Gradient-based Neural DAG Learning(GraN-DAG\cite{gran2020}, Learning Sparse Nonparametric DAGs(sparse-DAG\cite{sparse2020}), NOTEARS\cite{zheng2018}, NOTEARS nonlinear extension DAG-GNN\cite{yu2019} and causal additive models(CAM\cite{cam2014}). Comparisons with other methods like greedy equivalence search(FGS\cite{Chickering2002}) and PC\cite{Spirtes2000} omitted due to their week performance in recently research\cite{gran2020}\cite{sparse2020}\cite{yu2019}. We report the result of ture positive rate(TPR) and structural Hamming distance(SHD) as the evaluation metrics.      

\subsection{Synthetic Data}
We use Erdos-Renyi(ER) as the generation scheme of graph and use ERx for x$d$ edges, and then generate the data from SEM $X_j = f_j(X_{pa(X_j)}) + z_j$ for all j in topological order on the given graph. Functions $f_j$ can be Gaussian Process with a unit bandwidth RBF kernal and independent sampled $\sigma_j^2$ or MLP methods
with mutually independent noise $\sigma_j^2$ or Additive models with Gaussian Processes. In our experiments, we use Gaussian Processes with unit independent Gaussian noise.

We compare the performance of ER1 and Er4 for 10 nodes and ER1 for 50 nodes and we omit Er4 of 50 nodes as the SHD in this dataset is too large and it is meaningless to compare. The results of the comparison are shown on table1. We can see that NF-DAG performs best at 10 nodes, while CAM leads ER1 in 50 nodes, but CAM performs poorly in Er4 with 10 nodes. From this we can see that CAM is not suitable for dense edges. We also tested 10 nodes with 45 edges, and CAM performed even worse (the averaged SHD is 31.2 in the case of 5 samples), which was far worse than NF-DAG (SHD averaged 23.5 in 5 samples) and GraN-DAG(SHD averaged 25.1 in 5 samples). We think the reason is that NF-DAG uses a modern mature normalizing flow model MAF to fit the data, and the model can also stack batch normalizing layer and other technologies to enhance the fitting ability of the model. NOTEARS and its nonlinear extension DAG-GNN do not perform well in this dataset, because NOTEARS can only deal with linear causality while DAG-GNN uses linear adjacency matrix multiplied by nonlinear function to fit data, but it only uses adjacency matrix to represent causality and SPARSE-DAG has the same issue either.

\begin{table}
\centering
\caption{Comparison of different methods on non-linear SEMs generated from Gaussian processes(GPs) with unit independent Gaussian noise. The lower the better for SHD and the higher the better for TPR.}
\begin{tabular}{lll|ll|ll|llll} 
\cline{1-9}
 & \multicolumn{2}{l}{ER1 with 10 nodes} & \multicolumn{2}{l}{ER4 with 10 nodes} & \multicolumn{2}{l}{ER1 with 50 nodes}  &  &   \\ 
\cline{2-9}
                  & SHD        & \multicolumn{1}{l}{TPR}  & SHD        & \multicolumn{1}{l}{TPR}  & SHD  & TPR                          &  &   \\ 
\cline{1-9}
DAG-MAF           & \textbf{1.3$\pm$2.3} & \textbf{0.91$\pm$0.26}               & \textbf{16.4$\pm$4.9} & \textbf{0.77$\pm$0.12}               & 18.6$\pm$6.2 & 0.77$\pm$0.10                            &  &   \\
GraN-DAG          & 2.4$\pm$2.2          & 0.85$\pm$0.13                        & 18.6$\pm$4.1          & 0.66$\pm$0.11                        & 15.1$\pm$7.7          & 0.79$\pm$0.05                                               &  &   \\
Sparse-DAG        & 3.6$\pm$2.7          & 0.82$\pm$0.22                        & 20.1$\pm$6.7          & 0.63$\pm$0.10                        & 20.9$\pm$5.9          & 0.73$\pm$0.06                                                &  &   \\
CAM               & 5.1$\pm$2.1          & 0.90$\pm$0.06                        & 20.8$\pm$1.6          & 0.61$\pm$0.08                        & \textbf{5.3$\pm$1.8}          & \textbf{0.95$\pm$0.01}                                              &  &   \\
NOTEARS           & 4.8$\pm$3.0          & 0.62$\pm$0.18                        & 35.2$\pm$2.7          & 0.16$\pm$0.04                        & 22.8$\pm$7.1          & 0.66$\pm$0.12                                                &  &   \\
DAG-GNN           & 7.0$\pm$3.5          & 0.51$\pm$0.26                        & 37.0$\pm$2.2          & 0.12$\pm$0.09                        & 33.4$\pm$7.4          & 0.44$\pm$0.10                                            &  &   \\

\cline{1-9}
\end{tabular}
\end{table}

\subsection{Visualization of Causal Structure Forming}
\begin{figure}
    \centering
    \begin{subfigure}[b]{0.3\textwidth}
        \includegraphics[width=\textwidth]{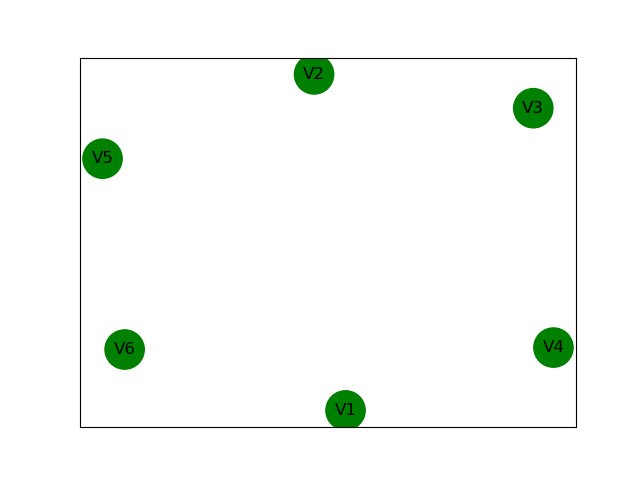} 
    \end{subfigure}
    \begin{subfigure}[b]{0.3\textwidth}
        \includegraphics[width=\textwidth]{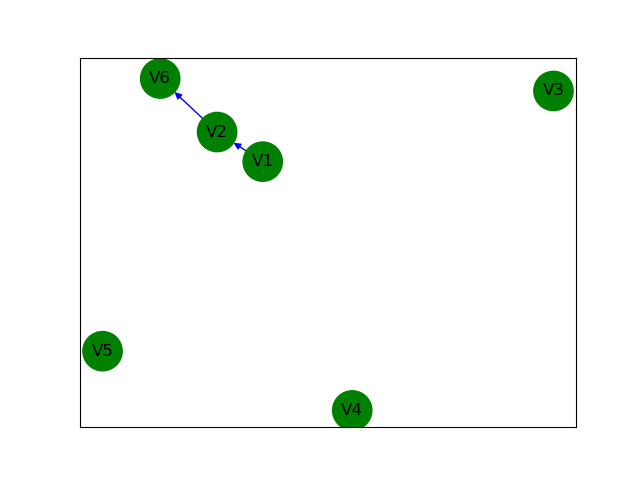}
    \end{subfigure}
    \begin{subfigure}[b]{0.3\textwidth}
        \includegraphics[width=\textwidth]{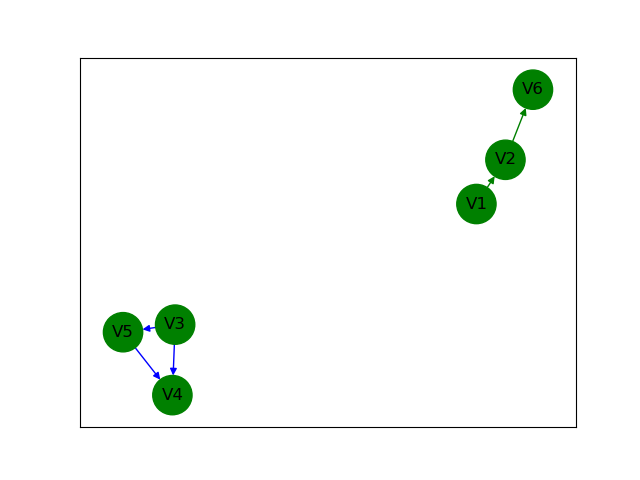}
    \end{subfigure}
    \begin{subfigure}[b]{0.3\textwidth}
        \centering
        \includegraphics[width=\textwidth]{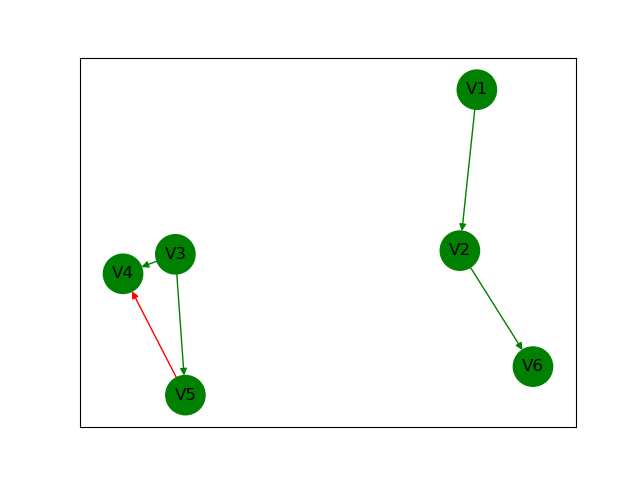}
    \end{subfigure}
    \begin{subfigure}[b]{0.3\textwidth}
        \centering
        \includegraphics[width=\textwidth]{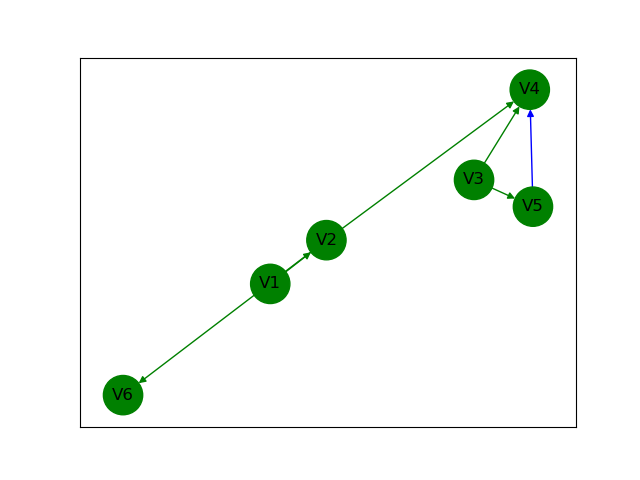}
    \end{subfigure}
    \begin{subfigure}[b]{0.3\textwidth}
        \centering
        \includegraphics[width=\textwidth]{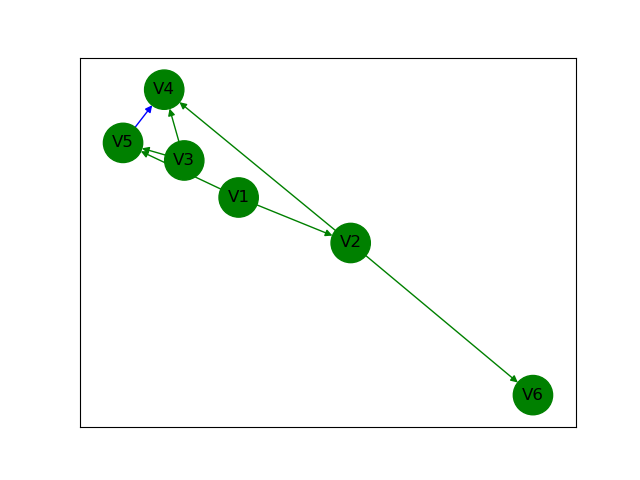}
    \end{subfigure}
    \begin{subfigure}[b]{0.3\textwidth}
        \centering
        \includegraphics[width=\textwidth]{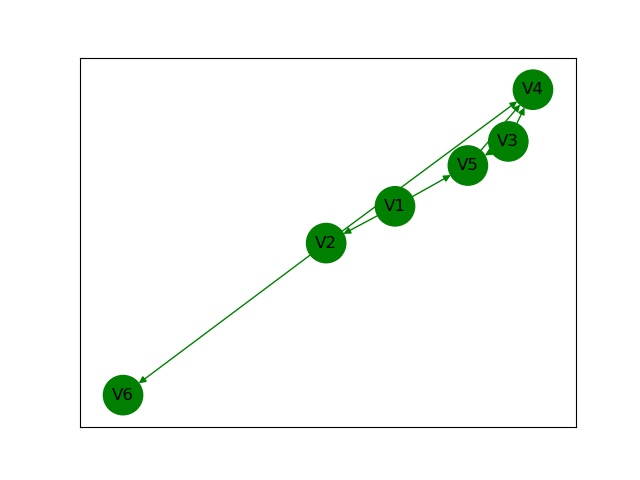}
    \end{subfigure}
    \caption{Visualization of the causal structure reconstruction from non-dependent Gaussian noises to the structure we learned by 6 blocks of flow-based generative models. The blues for new detected edges by new block of neural networks and the red ones for deleting edges from the last block learned.}
    \label{fig:vis1}
\end{figure}

We prepared a dataset with 6 variables from Additive models with Gaussian Processes and we can visualize the causal structure reconstructed from non-dependent Gaussian noises to the structure we expected. Please see the Figure \ref{fig:vis1} for the visualization. 

\subsection{Real Data}
\begin{figure}
    \centering
    \includegraphics[width=14cm]{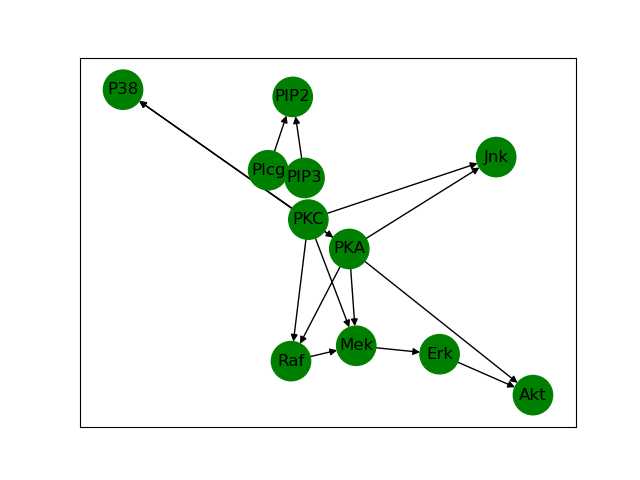}
    \caption{The ground truth causal graph of Sachs dataset}
    \label{fig:sachsgt}
\end{figure}

\begin{figure}
    \centering
    \includegraphics[width=14cm]{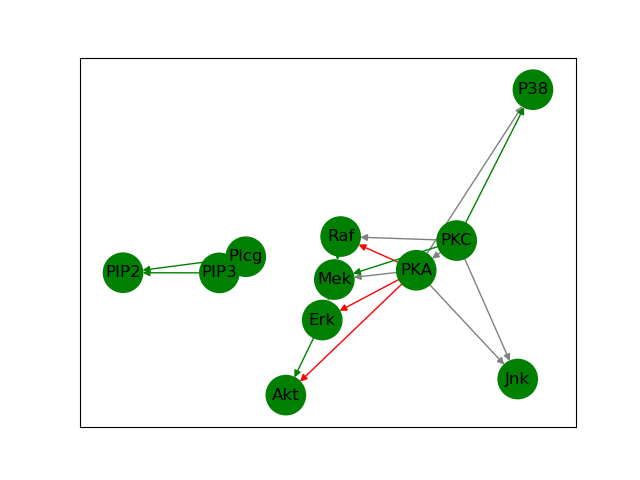}
    \caption{The causal structure learned by DAG-MAF: the green lines represent the expected edges; the gray ones for missing edges and the red ones for reversed edges.}
    \label{fig:sachs2}
\end{figure}

We evaluate the real dataset that is generally accepted by the biological community and is often used as a benchmark. The data consists of 11 continuous variables corresponding to different proteins and phospholipids in cells of the human immune system and 7466 observations, each of which indicates the measured level of each biological molecule in a single cell under different experimental interventions\cite{sachs2005}. The consensus network as causal graph is shown in Figure \ref{fig:sachsgt}. 
While the groundtruth of the The consensus network is 17 edges, we report SHD of 14 estimated 9 edges, in which there are 6 expected edges and 3 reversed as shown in Figure \ref{fig:sachs2}. For detail, the 6 true positives are Raf $\rightarrow$ Mek, Plcg $\rightarrow$ PIP2, PIP3 $\rightarrow$ PIP2, Erk $\rightarrow$ Akt, PKC $\rightarrow$ Mek and PKC $\rightarrow$ P38; the 3 reversed edges are PKA $\rightarrow$ Raf, PKA $\rightarrow$ Erk and PKA $\rightarrow$ Akt; 8 missing edges are Mek $\rightarrow$ Erk, Plcg $\rightarrow$ PIP3, PKA $\rightarrow$ Mek, PKA $\rightarrow$ P38, PKA $\rightarrow$ Jnk, PKC $\rightarrow$ Raf, PKC $\rightarrow$ PKA and PKC $\rightarrow$ Jnk. By comparison, while DAG-GNN reports SHD of 19 with 18 edges predicted, GraN-DAG estimated 16 edges with SHD of 13 and Sparce-DAG predicted 13 edges with SHD of 16.       

\section{Conclusion}
In this work, we propose a jacobian-based normalizing flow method called DAG-MAF to learn dependencies of input observation data. Inspired by Grad-CAM in computer vision, we use jacobian matrix of output w.r.t. input as causal relationships and this method can be generalized to any neural networks especially for flow-based generative neural networks such as Masked Autoregressive Flow(MAF) which compute the log likelihood loss and divergence of distribution of input data and target distribution. This method extends NOTEARS which enforces a important acylicity constraint on continuous adjacency matrix of graph nodes and significantly reduce the computational complexity of search space of graph. We did massive experiments and the results show that our method outperform the original NOTEARS, its nonlinear extension DAG-GNN and other machine learning based methods such as GraN-DAG.

\bibliographystyle{unsrt}  


\end{document}